\documentclass{article}
\usepackage{spconf,amsmath,graphicx}

\title{Estimating exercise-induced fatigue from thermal facial images}

\name{Manuel Lage Cañellas 
\qquad Constantino Álvarez Casado 
\qquad Le Nguyen 
\qquad Miguel Bordallo López }
\address{Center for Machine Vision and Signal Analysis (CMVS), University of Oulu, Finland}
       
\begin{document}
\maketitle

\begin{abstract}
Exercise-induced fatigue resulting from physical activity can be an early indicator of overtraining, illness, or other health issues.
In this article, we present an automated method for estimating exercise-induced fatigue levels through the use of thermal imaging and facial analysis techniques utilizing deep learning models.
Leveraging a novel dataset comprising over 400,000 thermal facial images of rested and fatigued users, our results suggest that exercise-induced fatigue levels could be predicted with only one static thermal frame with an average error smaller than 15\%. The results emphasize the viability of using thermal imaging in conjunction with deep learning for reliable exercise-induced fatigue estimation.

\end{abstract}


\section{Introduction}
\label{sec:intro}

Exercise-induced fatigue refers to a strong sensation of exhaustion that occurs due to physical exertion.
When engaging in physical activity, the human body raises thermal radiation, which is closely tied to the experienced fatigue.
In this context, thermal cameras, designed to measure emitted heat, demonstrate remarkable effectiveness in capturing fluctuations in skin temperature associated with fatigue.
This distinctive capability makes thermal cameras particularly valuable for detecting exercise-induced fatigue without the need for physical contact or causing disruptions to the environment.


In this work, we propose the estimation of the level of exercise-induced fatigue using thermal images obtained from healthy individuals using facial analysis and deep learning. Our contributions can be summarized as follows:
\begin{itemize}
\setlength\itemsep{0pt}
\setlength\parskip{0pt}
\item We introduce for the first time the use of thermal facial images to estimate the intensity of exercise-induced fatigue in healthy people.
\item We leverage a novel dataset of 418,813 thermal images from 80 subjects that we annotate with fatigue levels ranging from 0 (resting) to 100 (heavily fatigued).
\item We show that a regression of the level of fatigue can be performed from individual facial images. Employing residual deep convolutional neural networks (ResNet), we obtain an average error of 13.66\% for the best case with a standard deviation among users of 14.43\%.
\item Additionally, we examine the results stratified by gender and facial accessory (glasses) and address the limitations of our labelling annotation system.
\end{itemize}


\subsection{Related work}

Exercise-induced fatigue assessment has traditionally encountered issues related to invasiveness, inaccuracy, and practical limitations. For instance, the Rating of Perceived Exertion (RPE), a subjective measure of exercise intensity, employs the Borg RPE scale, where individuals self-assess their exertion on a scale of 0-10 or 6-20 \cite{ekblom1971influence}. However, this method is susceptible to biases from mood, motivation, and expectations, hindering result comparability and longitudinal tracking, as underscored by Lamb et al. \cite{lamb1999reliability}.

The rise of computer vision has introduced non-intrusive and non-wearable fatigue detection methods, addressing these limitations. For example, in \cite{irtija2018fatigue}, the authors propose a fatigue detection technique leveraging alterations in the eye and mouth regions based on facial landmark points. This involves computing a central point from 68 detected landmark coordinates and measuring distances between eye and lip corners, along with eye width. These metrics serve as features input to a Support Vector Machine (SVM) classifier, discerning signs of fatigue.
Another illustration comes from \cite{uchida2018identification}, where the Facial Action Coding System (FACS), utilized by psychologists to identify emotional facial expression patterns, is combined with facial muscle activation (sEMG) during exercise at varying intensities. This aims to uncover connections between facial expression shifts, exercise intensity, and fatigue levels, thereby exploring potential correlations.
In a similar vein, Haque et al. \cite{haque2016facial} exemplify the utilization of computer vision techniques. By extracting trajectories from facial points through the Good Features to Track (GFT) approach, these trajectories are further tracked using a supervised descent method to estimate physical fatigue.

Computer vision techniques based on RGB cameras circumvent the limitations of conventional fatigue assessment approaches, enabling the exploration of relationships between facial expressions, and fatigue levels during exercise. Thermal imaging provides a potentially interesting alternative that could also measure muscle activity and heat exchange patterns, but has been infrequently employed in facial analysis \cite{krivsto2018overview}, especially when related to affective computing and fatigue. Among the limited studies, to the best of our knowledge only the work by \cite{lopez2017detecting} specifically addressed the detection of exercise-induced fatigue, treating it as a binary classification problem, using a modest dataset. We extend this research, aiming to determine instead the specific levels of fatigue in individuals with a larger dataset.

\section{Methodology}

\subsection{Benchmark data}

In a similar fashion as previous work \cite{lopez2017detecting}, we collect a dataset consisting of 160 different videos using a thermal camera to calculate the fatigue level based on two different conditions: fatigued and rested individuals. The data was obtained from 80 participants (51 male, 29 female) between 18 and 68 years old. Each participant recorded two video sessions, shooting for about five minutes. In the first session, participants were recorded in resting condition, meaning that their heart and respiratory rates were below 80 bpm and 12 rpm, respectively. For the second session, fatigue was induced by intense exercise, starting the recording only if the heart rate was above 120 bpm and the respiratory rate above 15 rpm. Physical exertion ensured a heart rate between 60\% and 80\% of maximum heart rate (HRM) calculated with the common HRM = 220-age formula \cite{nes2013age}. During the second session, subjects slowly recovered from the fatigue, which we ensured by checking that the video ended with participants in previously described resting condition.

Resting videos were assigned a fatigue level of 0, while those recorded post-exercise were assigned decreasing fatigue levels that diminished from 100 to 0 as the video progressed. 
Although relatively arbitrary, this recovery pattern coincides with the almost linear recuperation of phosphocreatine, a molecule related with muscle fatigue \cite{sahlin1998energy}, during the first five minutes of rest after physical exercise \cite{roussel200031p}, providing a fine-grained measure of the fatigue experienced by the individuals.

Thermal cameras detect emitted infrared radiation, but glasses, often made from transparent materials, can absorb thermal radiation rendering glass opaque to thermal cameras, causing partial occlusion of faces. 
To address this issue, we ensured the proper stratification of the dataset, maintaining an equal distribution of gender and occlusion across the different folds using in cross-validation.

\begin{figure}[t]
 \begin{center}
   \includegraphics*[width=0.9\columnwidth]{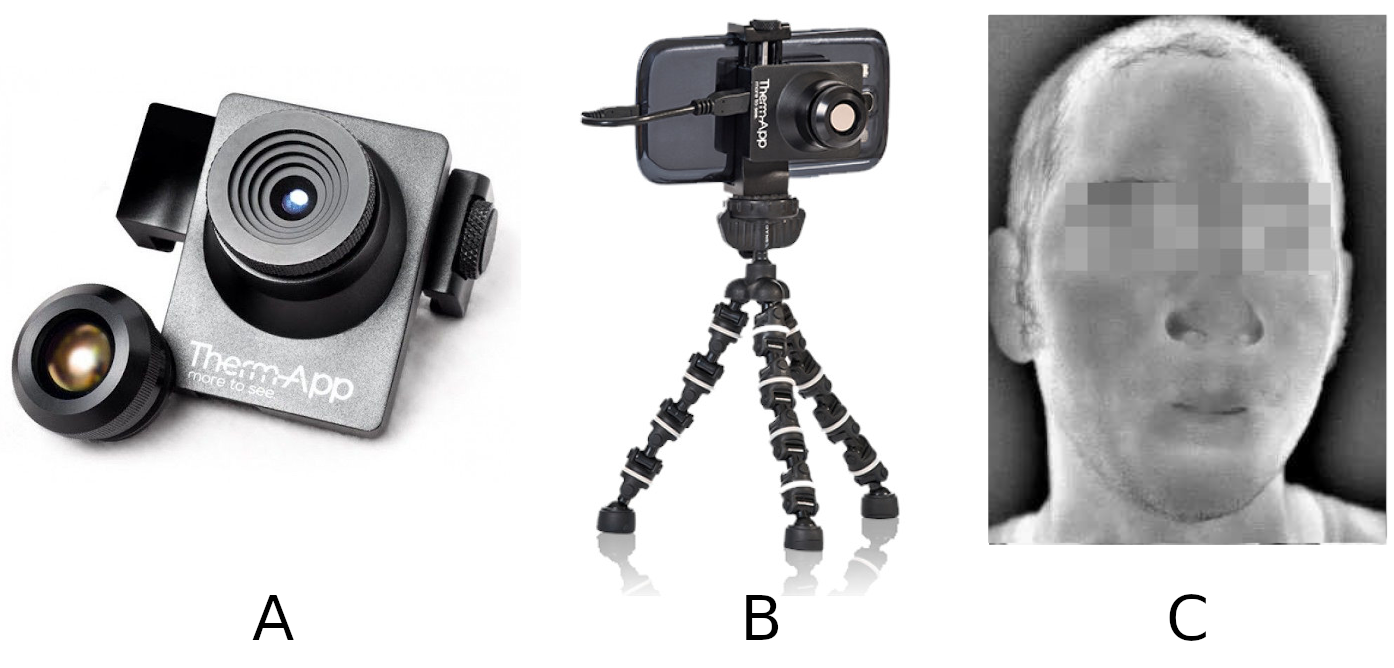}
 \end{center}
 \vspace{-7mm}
 \caption{A: Therm-App mobile thermal camera. B: Therm-App camera mounted in an android phone. C: Example of a frame captured with Therm-App thermal.}
 \label{fig:thermalcamera}
\end{figure}

The facial videos were acquired utilizing the Therm-App camera installed on a tripod as shown in Figure \ref{fig:thermalcamera} B). 
The thermal camera features Very Long-Wave Infrared (VLWIR) 17 {\textmu}m thermal detector, and is equipped with a 19mm lens with manual focus providing a resolution of 288 by 384 pixels at 8.7Hz.
The thermal data range was compressed to 256 levels, represented in grayscale. Additionally, the temperature dynamic range was clipped, such that the mean temperature of the image corresponds to level 128, while levels 0 and 255 correspond to 90\% and 10\% of the maximum. 
The facial images were captured at a distance of approximately one meter from the subject, which maximizes the resolution of the face for the chosen lens.

\subsection{Preprocesssing}
Our initial experiments revealed that face detection algorithms, such as Multi-task Cascade Convolutional Network (MTCNN) \cite{ZhangMtcnn2016}, exhibited suboptimal performance when applied to thermal images, a largely known problem \cite{poster2019examination}. 
Since our dataset was collected under consistent conditions, with subjects positioned at a fixed distance from the camera, 
a minimum interpupillary distance of 100 pixels \cite {lopez2017detecting} was preserved.
Hence, our preprocessing consists in cropping a central region of the image.
Since the subjects were asked not to move during the recording and the camera was set to have the face approximately in the center of the image, their positions tend to be consistent across all frames, significantly reducing the need for additional alignment. We resize the images to match the input of the neural networks, and added data augmentation by using random 50\% horizontal flip. 


\subsection{Deep Learning Models}
\begin{figure}[t]
 \begin{center}
   \includegraphics*[width=1.0\columnwidth]{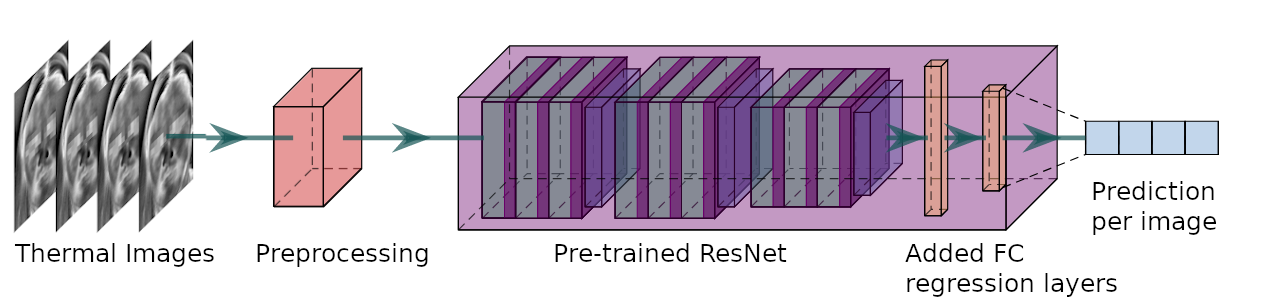}
 \end{center}
 \vspace{-7mm}
 \caption{Modified ResNet architecture. Two fully connected layers were added to regress the fatigue levels.}
 \label{fig:architecture}
\end{figure}

The experiments in this study involved the use of various combinations of ResNet architectures \cite{resnet2016}
as the backbone, as they have been proven useful to other similar tasks such as driving fatigue detection \cite{sikander2020novel}.
To regress the level of fatigue, two fully connected (FC) layers were added to the original ResNet, as depicted in Figure \ref{fig:architecture}. The first added layer varies in size depending of the original ResNet structure whereas the second layer is a regression layer with fixed size of 128. To take advantage of pre-learnt features and transfer learning, different pre-trained weights were utilized, such as \textit{IMAGENET\_1K\_V2} based on Imagenet \cite{deng2009imagenet} tested on several ResNet architectures, as well as weights based on the InsightFace (InsightResNet) framework \cite{guo2021sample}. All layers of the models were left unfrozen at a small learning rate to allow the retention of low-level textural features. Finally, each inference shows the prediction of one video frame.

\subsection{Training and evaluation}
The thermal dataset contains a total of 418,813 frames. We employed a five-fold cross-validation to evaluate the performance. The cost of this regression model is calculated with the L1 norm. We include RAdam \cite{liuRadam2019} optimizer, enhanced with an implementation of Lookahead \cite{LookaheadZhang2019} and a ReduceLROnPlateau scheduler with an initial learning rate of $3\times10^{-4}$ \cite{BermantBioCPPNet2021}. To evaluate the performance of our models, we provide the two most common metrics for regression problems, mean absolute error (MAE) and root mean squared error (RMSE). This pretrained scheme has shown to be useful for a similar regression task with textural information \cite{lage2022improving}.

\section{Results}
We train a set of different models based on ResNet architectures with different number of layers, ranging from 18 to 101 layers, and different pretraining schemes, including ImageNet and Insight Face. The MAE and RMSE results obtained for the models are presented in Table \ref{tab:sota}.

\begin{table}[ht!]
 \vspace{-1mm}
\setlength{\tabcolsep}{0.75em}
\def\arraystretch{1.1}
\begin{center}
\begin{tabular}{|l|c|c|}
\hline
Models & MAE ± std & RMSE ± std \\
\hline
Insight-ResNet 50  & 16.18 ± 15.58 & 20.19 ± 17.83 \\
Insight-ResNet 100 & 15.74 ± 15.50 & 19.82 ± 17.84 \\
ImageNet-ResNet 34         & 15.54 ± 14.44 & 18.93 ± 16.47 \\
ImageNet-ResNet 101        & 15.48 ± 15.54 & 18.99 ± 17.31 \\
ImageNet-ResNet 18         & 15.37 ± 14.83 & 18.64 ± 16.75 \\
Insight-ResNet 34  & 15.10 ± 14.54 & 18.56 ± 16.34 \\
ImageNet-ResNet 50         & 13.64 ± 14.43 & 16.56 ± 16.17 \\
\hline
\end{tabular}
\end{center}
\vspace{-5mm}
\caption{Comparison of results of exercise-induced fatigue results for different ResNet architectures on our dataset.}
\label{tab:sota}
\end{table}

By observing the results we can see that the MAE is similar across different models with a relatively small 3\% variation among them. No particular pre-training strategy seems to be consistently better, and given the small size of the datasize, slightly good results can be obtained with small network sizes.

\begin{figure}[t]
 \begin{center}
   \includegraphics*[width=0.5\columnwidth]{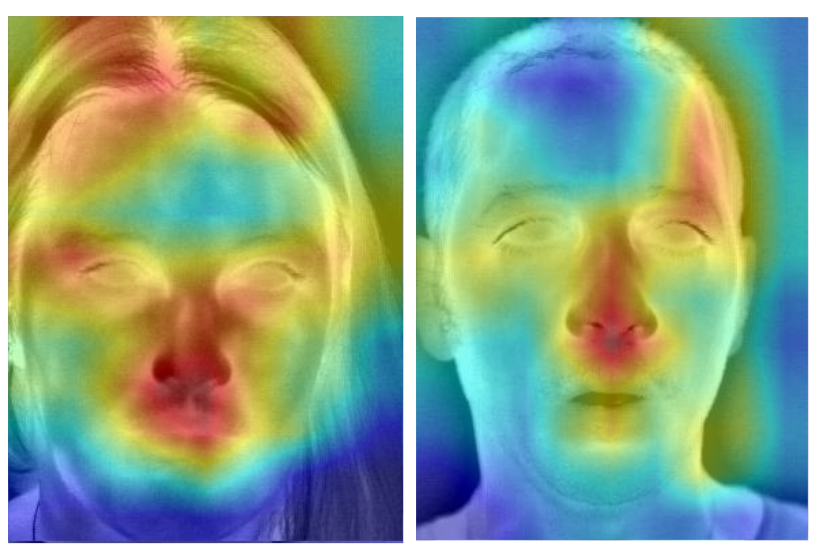}
 \end{center}
 \vspace{-7mm}
 \caption{Camera Grad-CAM applied to ImageNet-ResNet 50 model shows in red the neural network regions of interest.}
 \label{fig:grad_cam}
\end{figure}

In order to determine which parts of the face contribute most significantly to the ResNet regression level, we utilized a Grad-CAM (Gradient-weighted Class Activation Mapping) to determined those regions of interest considered important by the model \cite{jacobgilpytorchcam}. As seen in Figure \ref{fig:grad_cam}, the example Grad-CAM maps indicate that the ROIs belong mostly the nose and mouth regions of the face, which are the areas of heat exchange in the human face. 

\begin{figure}[ht!]
 \vspace{-2mm}
 \begin{center}
   \includegraphics*[width=0.47\textwidth]{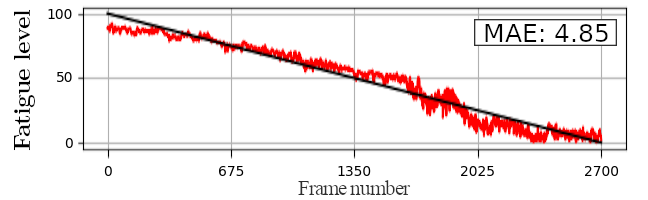}
 \end{center}
 \vspace{-8mm}
 \caption{Individual fatigue predictions (red) against our labelling (black). Case of consistent fatigue-decay correlation.}
 \label{fig:A1}
 \vspace{-3mm}
\end{figure}

The predictions of individual frames for a fatigued video of one user are illustrated in Figure \ref{fig:A1}.
The graph displays a clear correlation between individual frames and fatigue levels. 
This is an intriguing finding as it suggests that an individual's level of fatigue could be predicted with only one static frame. 
The results suggest that there is a strong correlation between the predicted values and the rate of fatigue decay.
In addition, we observed that among the users with the highest error, two distinct classes could be distinguished: misclassified users and users with a decay ratio different from the one arbitrarily assigned during labeling. 

\begin{figure}[ht!]
 \begin{center}
   \includegraphics*[width=0.47\textwidth]{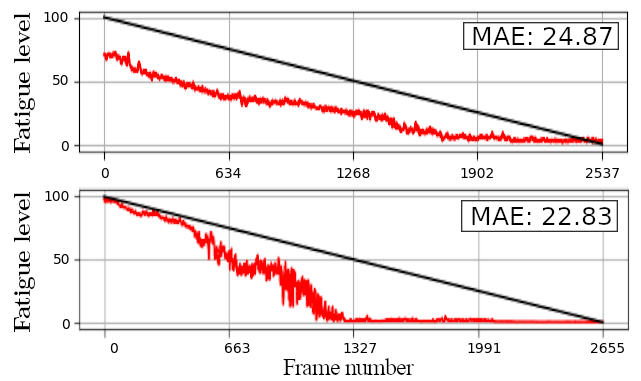}
 \end{center}
 \vspace{-7mm}
 \caption{The top subplot shows an individual that probably started with a smaller level of fatigue,  while the bottom subplot belongs to an individual with a fast recovery time.}
 \label{fig:A2}
\end{figure}

Figure \ref{fig:A2} shows examples of individuals with high regression error, but that still show a clear correlation between the fatigue level prediction and the label. This observation points out that the decay ratio for these users might have been different from the one assigned during labeling, suggesting that the individuals may have started the test with a lower fatigue level than expected, or that they were able to recover faster than normal. Although this indicates that arbitrarily labelling the fatigue levels from 100 to 0 might be a sub-optimal approach, the predicted levels of fatigue still decayed gradually and consistently throughout the video. These observations further highlight the importance of individual-level analysis in developing fatigue prediction models that cater to individual characteristics and needs, and it is left for future work.

Table \ref{tab:mae_strat} examines the influence of gender and glasses on the model's accuracy. Our findings suggest that the presence of glasses affects the detection of rested states more than fatigue states. This might cause rested users, especially in high-temperature change areas like the mouth and nose, to appear more fatigued. When glasses are worn during rest, the diminished temperature variation and associated labels might decrease the detail captured by thermal cameras. However, our analysis revealed an equal distribution of errors across stratified groups. Although glasses appear to affect predictions for resting states, this effect is likely attributed to a few outliers rather than a systemic bias. In terms of gender differences, our study found almost none, pointing to a consistent performance of the model across genders. In addition, the study outcomes show a balanced distribution between fatigue and resting level estimations.

\begin{table}[ht!]
 \vspace{-1mm}
\setlength{\tabcolsep}{0.45em}
\def\arraystretch{1.0}
\begin{center}
\begin{tabular}{|l|c|c|c|c|}
\hline
Group & Combined &Fatigue &Resting \\
\hline
Men + Women   & 13.64 & 22.20 & 5.40 \\
Men   & 13.46 & 23.72 & 3.59 \\
Women & 13.96 & 19.52 & 8.60 \\
\hline
Men + Women no glasses   & 14.01 & 21.44 & 6.57 \\
Men no glasses   & 13.77 & 21.97 & 5.57 \\
Women no glasses & 14.32 & 20.74 & 7.91 \\
\hline
Men + Women with glasses  & 13.03 & 23.52 & 3.56\\
Men with glasses  & 13.06 & 26.07 & 1.18\\
Women with glasses& 12.96 & 15.88 & 10.40\\
\hline
\end{tabular}
\end{center}
\vspace{-5mm}
\caption{Mean Average Error results stratified by gender and glasses}
\label{tab:mae_strat}
\end{table}

\begin{figure}[t]
 \begin{center}
   \includegraphics*[width=0.45\textwidth]{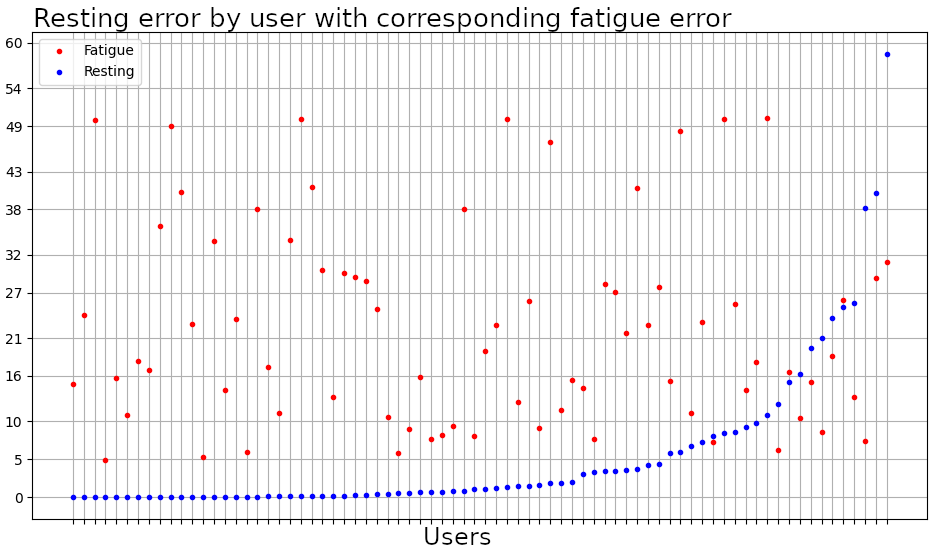}
 \end{center}
 \vspace{-5mm}
 \caption{Sorted MAE by user Stratified results by user. No correlation between errors in resting and fatigue for the same user.}
 \label{fig:stratification_2}
 \vspace{-5mm}
\end{figure}

We present an analysis of the model's performance, differentiated by fatigue and rested states, as illustrated in Figure \ref{fig:stratification_2}. In this figure, each vertical line represents an individual subject. The red and blue dots correspond to the error rates for the fatigue and rested conditions, respectively. Subjects are ordered based on their error rates in the rested state. Notably, our analysis indicates that a higher error rate in one state does not correlate with a higher error rate in the other.

The absence of a strong correlation suggests that the model may not be discerning broader contextual cues that differentiate the two states but is instead focused on more localized patterns. 
This implies that the model might be honing in on textural features, potentially driven by the unique thermal emission patterns of individuals.
This observation aligns with visualizations presented in the Grad-CAM \ref{fig:thermalcamera}.

The present study investigates the correlation between thermal camera frames and levels of fatigue induced by exercise. 
Although our model seems to grasp this connection, we acknowledge the limitations of our existing labeling system, which neglects the individual fatigue decay ratio.
Our discoveries underscore the significance of conducting a thorough assessment of the model's performance across different states and propose the need for advancing more reliable fatigue prediction models.

\section{Conclusion}
This research presents a novel system for estimating exercise-induced fatigue levels, based on individual thermal images. 
Our procedure involves the use of thermal cameras that provide a reliable, illumination-invariant technique.
The results of our method using a new dataset of 418,813 thermal images from 80 subjects, suggest the feasibility of accurately determining fatigue levels from thermal images, making a step forward in fatigue management.
Our stratified study of the dataset allowed us to determine the reliability of different methods for estimating fatigue levels.
We found that residual neural networks offer a particularly consistent framework that provides low bias based on gender and facial accessories such as glasses.
This finding underscores the potential of deep learning methods for accurately estimating fatigue levels from thermal images.
While our research focuses solely on static images, there is significant scope for future studies to explore the complementary nature of diverse sources by utilizing various biosignals.
Additionally, future work could address the creation of labels based on biosignals such as heart rate and respiration rate to overcome the shortcomings of the different fatigue ratio decays among users.
In conclusion, our work has demonstrated the potential of thermal imaging and deep learning methods for accurately estimating exercise-induced fatigue levels from individual thermal images.

\bibliographystyle{IEEEbib}
\bibliography{strings,refs}

\end{document}